\documentclass[conference]{IEEEtran}
\IEEEoverridecommandlockouts
\usepackage{cite}
\usepackage{amsmath, amssymb, amsfonts}
\usepackage{algorithmic}
\usepackage{textcomp}
\usepackage{xcolor}
\usepackage{url}
\usepackage{hyperref}

\def\BibTeX{{\rm B\kern-.05em{\sc i\kern-.025em b}\kern-.08em
    T\kern-.1667em\lower.7ex\hbox{E}\kern-.125emX}}

\usepackage[english]{babel}
\usepackage[utf8]{inputenc}
\usepackage[T1]{fontenc}
\usepackage{lmodern}

\usepackage{isomath}
\usepackage{amsmath}

\usepackage{graphicx}
\usepackage{color}
\usepackage{epstopdf}
\graphicspath{{figures/}}
\begin{document}
\title{Omnidirectional robot modeling and simulation
\thanks{This work is financed by National Funds through the Portuguese funding agency, FCT - Fundação para a Ciência e a Tecnologia, within project UIDB/50014/2020, and the ERDF – European Regional Development Fund through the Operational Programme for Competitiveness and Internationalisation - COMPETE 2020 Programme, and by National Funds through the Portuguese funding agency, FCT - Fundação para a Ciência e a Tecnologia, within project SAICTPAC/0034/2015- POCI-01-0145-FEDER-016418. \\ The final publication is available at IEEE via \url{http://dx.doi.org/10.1109/icarsc49921.2020.9096069}.}}

\author{\IEEEauthorblockN{Sandro A. Magalhães}
\IEEEauthorblockA{\textit{INESC TEC} \\
\textit{Faculty of Engineering,}\\
\textit{University of Porto} \\
Porto, Portugal \\
orcid.org/0000-0002-3095-197X}
\and
\IEEEauthorblockN{António Paulo Moreira}
\IEEEauthorblockA{\textit{INESC TEC} \\
\textit{Faculty of Engineering,}\\
\textit{University of Porto} \\
Porto, Portugal \\
orcid.org/0000-0001-8573-3147}
\and
\IEEEauthorblockN{Paulo Costa}
\IEEEauthorblockA{\textit{INESC TEC} \\
\textit{Faculty of Engineering,}\\
\textit{University of Porto} \\
Porto, Portugal \\
orcid.org/0000-0002-4846-271X}
}

%
%
%
%
\maketitle              
\begin{abstract}
A robots simulation system is a basis need for any robotics application. With it, developers teams of robots can test their algorithms and make initial calibrations without risk of damage to the real robots, assuring safety. However, build these simulation environments is usually a time-consuming work, and when considering robot fleets, the simulation reveals to be computing expensive.  With it, developers building teams of robots can test their algorithms and make initial calibrations without risk of damage to the real robots, assuring safety. An omnidirectional robot from the 5DPO robotics soccer team served to test this approach. The modeling issue was divided into two steps: modeling the motor's non-linear features and modeling the general behavior of the robot. A proper fitting of the robot was reached, considering the velocity robot's response.

\end{abstract}


\section{Introduction}
\label{sec:intro}

Mobile robotics is waking up a high-level interest in different applications such as military issues, search and rescue, cleaning, inspection, or even some research areas, such as robotics soccer, whose researchers use to test new approaches or algorithms performance.

However, although mobile robots reduce their complexity to three degrees of freedom (DoF), which is smaller when compared with most of the manipulators, the control problem continues a high-demanding task~\cite{siegwart}. Some environments could become particularly complexes and dynamics, filled with robot fleets or other moving bodies that could not share information, increasing the environment unpredictability.

Mobile robots may have different kinds of traction. The most commons are differential traction, tricycle, and omnidirectional. The omnidirectional robots offer a higher degree of freedom and could present a topology of three or four wheels (rarely more) or MECANUM~\cite{siegwart,Tang}.

All robots have got to be well-calibrated before being used in real situations because of their unique features related to kinematics and dynamics constraints. Examples of these calibration parameters are the robots' controllers, such as the wheels speed controllers, and the position or trajectory controller. However, these calibration issues could be dangerous due to the system instability region, which could still be unknown and time-expensive to acquire. Therefore an initial simulation approach to calibrate these parameters and preview the instability regions before testing the algorithms into a real robot looks easier and safer.

Otherwise, finding a full simulation model of a robot is hard, and its usage is computationally intensive~\cite{ZLAJPAH}, bringing no simulation advantages, once the computer cannot compute the simulation model in real-time.

\begin{figure}
    \leavevmode
    \begin{center}        
        \includegraphics[height=5cm]{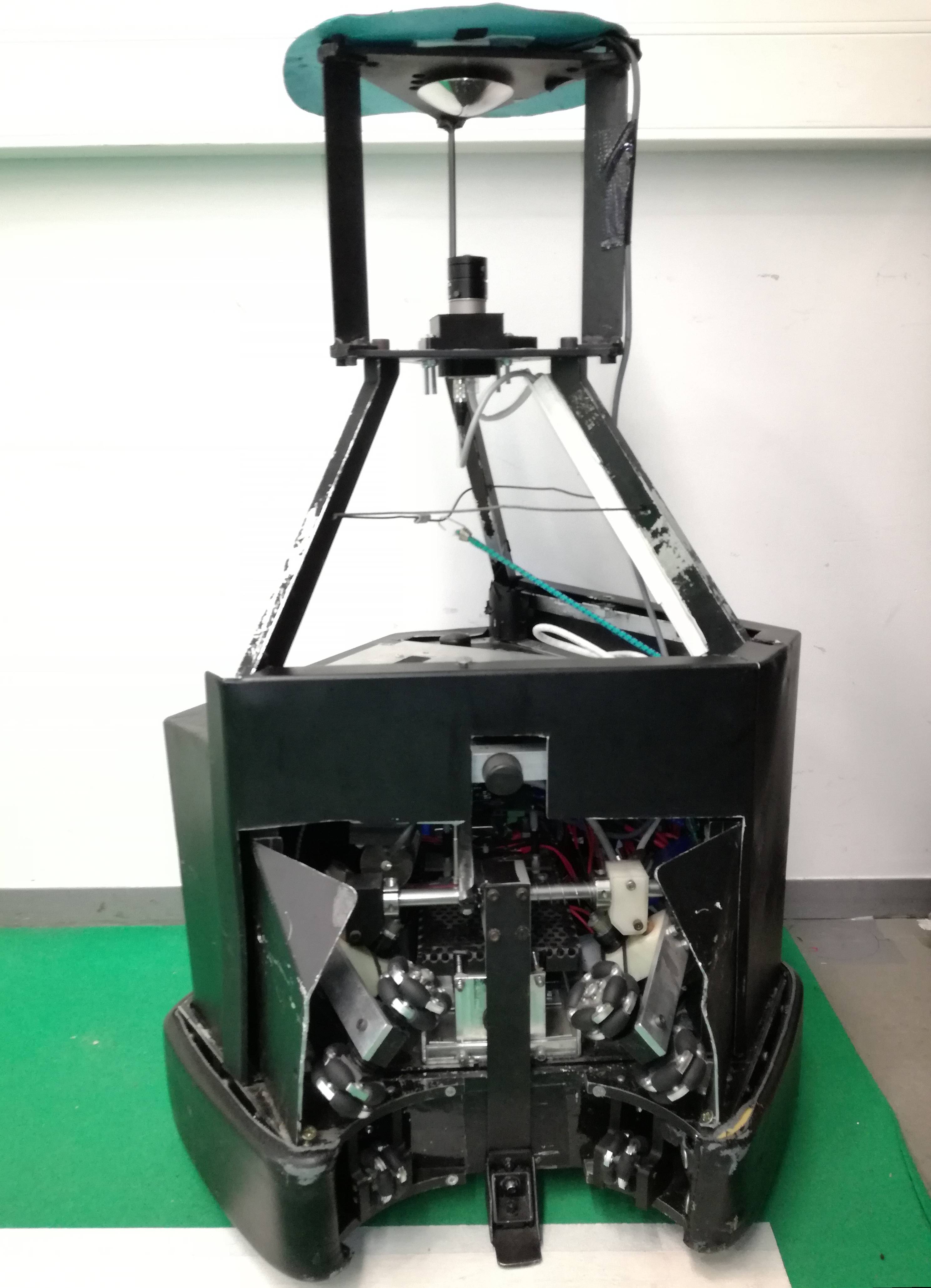}        
        \label{fig:5DPO_robot_soccer}
        \caption{Robot from 5DPO robotics soccer team}
    \end{center}
\end{figure}

Therefore, this paper intends to report an easier robot modeling strategy for simulation issues, using iterative algorithms such as the steepest descendant\cite{Yuan} or RPROP\cite{Riedmiller1993}, basing on real robotics behavior samples. This case study was applied to an omnidirectional robot from robotics soccer 5DPO team~\cite{5dpo}, illustrated on Fig.~\ref{fig:5DPO_robot_soccer}.

Section~\ref{sec:related_work} references the previously done work to develop a simulation model of the robots of the 5DPO robotics soccer team. After, section~\ref{sec:model_components} divides the modeling issues into two main components and describe them theoretically. The following section describes the modeling strategy, and section~\ref{sec:results} discusses the results. Finally, section~\ref{sec:conclusions} does an overview of the done work and leaves some future work for better validation of the gotten results.

\section{Related work}
\label{sec:related_work}

Other authors did previous works to model older versions of robots of the 5DPO robotics soccer team to different issues, namely, to simulate, to test and to design different kinds of algorithms in a controlled environment~\cite{Caldas2009}, and to design robot's controllers~\cite{Conceicao2007,Nascimento2009}.

Conceição \textit{et al.}~\cite{Conceicao2006,Conceicao2007,Conceicao2009} modeled a four-wheeled omnidirectional robot. To reach the full model of the robot, they divided this problem in two main steps, estimate the linear model of the motor and model the kinematic and dynamic model of the robot. In the first step, they approximated the motor's model using techniques based on the minimum square estimator and the variation of the instrumental variable~\cite{Conceicao2007}. In the second step, they estimated the non-linear features of the robots, using three different but complementary strategies: (i) estimation considering the velocity in steady-state, (ii) null traction, and (iii) the moment of inertia.
However, although they got a full and precise model of the robot with these approaches, the resulting model is very complex and computing demanding to simulate, besides the high effort and many measures needed in order to obtain the model.

Later, Caldas~\cite{Caldas2009} assembled a three-wheeled omnidirectional robot for simulation in the SimTwo simulator~\cite{SimTwo}, using a descriptive model to characterize the real robot features. This approach is characteristic of the simulator, and the robot's features are described into an XML file, using descriptive tags. For some specific parameters, such as the moment of inertia, controller's gains, and friction, he used a similar approach to Conceição \textit{et al}~\cite{Conceicao2006,Conceicao2007,Conceicao2009}. Because of his complete approach to characterize the robot, the simulation was too compute-intensive, and good results were hard to reach.

Besides the SimTwo, there are other robotics simulators more commonly used. Examples of them are Gazebo~\cite{gazebo}, or V-REP (now formally CoppeliaSim)~\cite{vrep}. However, SimTwo~\cite{SimTwo} is a free and open-source\footnote{\url{https://github.com/P33a/SimTwo}} robotics simulation software. It models the robots using descriptive tags in an XML file, which allows easy robots and environment design, and accessible parameters and controllers calibration. Each robot is composed of physical bodies interconnected through joints, which can have electrical motors with well-defined controllers. ODE library~\cite{Smith2007} powers the whole simulation environment that can communicate with external plugins to reach high-level decision algorithms or artificial intelligence, or execute simulator embedded scripts.

This work intends to create a simulation model for SimTwo, using Caldas' principles~\cite{Caldas2009} and a simplified methodology to parametrize robot dynamics for simulation.

\section{Model components}
\label{sec:model_components}

It is possible to divide the architecture of any robot into two main components: the kinematic model and the dynamic model. The kinematic model tries to characterize the robot through its constructive particularities, usually, the type and the geometry of its traction system. Otherwise, the dynamic model characterizes the robot's physics properties through differential equations. Some examples of these properties are the effect of inertia and the friction in the robot behavior.

For small robots that move slowly, the usage of just their kinematics models usually is sufficient. However, for more heavy robots that move fast, the dynamics complicate the control of the robots. So, for the situation of the Medium Size League (MSL) robotics soccer robots\footnote{https://msl.robocup.org/}, the simulation should consider the robots' dynamics.

\begin{figure}
    \leavevmode
    \begin{center}        
        \def\svgwidth{0.9\linewidth}
        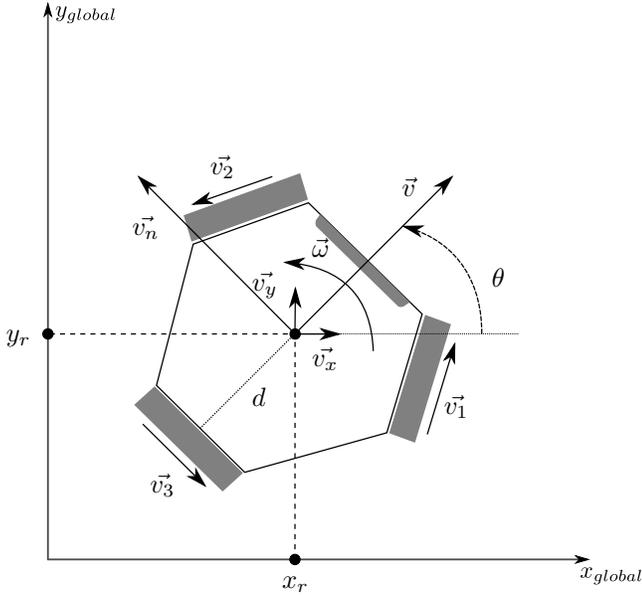
        \caption{Three wheels omnidirectional robot schematics. $\Vec{v}$, $\Vec{v_{n}}$ and $\Vec{\omega}$ are the linear, linear orthogonal and angular velocities of the robot, $v_{1}$, $v_{2}$ and $v_{3}$ are the linear velocities of the wheels and $d$ is the distance between the wheels and the geometric centre of the robot. $x_{r}$, $y_{r}$ and $\theta$ are the positions of the robot relative to the global frame.}
        \label{fig:robotSchemes}
    \end{center}
\end{figure}

The mobile robot identified on Fig.~\ref{fig:5DPO_robot_soccer}, and schematically represented of Fig.~\ref{fig:robotSchemes}, is a three-wheeled omnidirectional robot. The wheels are distributed in $120^{o}$ spacing angles and distanced of 19.5~cm $(d)$ from the center of the robot. Considering these features, the kinematic model of the robot is the conversion between the robot's velocity to the wheels' velocity, as illustrated in \eqref{eq:kinematics}, that refers to Fig.~\ref{fig:robotSchemes}, where $v_{1}$, $v_{2}$, and $v_{3}$ are the instantaneous velocities of the three omnidirectional wheels, $v$, $v_n$, and $\omega$ are the norm of the vectors represented on Fig.~\ref{fig:robotSchemes} and $d$ is the shorter distance between the wheel and the geometric center of the robot.

\begin{equation}
    \begin{bmatrix}
        v_1 \\ v_2 \\ v_3
    \end{bmatrix} = 
    \begin{bmatrix}
        \sin(\frac{\pi}{3}) & \cos(\frac{\pi}{3}) & d \\
        - \sin(\frac{\pi}{3}) & \cos(\frac{\pi}{3}) & d \\
        0 & -1 & d
    \end{bmatrix} \cdot
    \begin{bmatrix}
        v \\ v_n \\ \omega
    \end{bmatrix}
    \label{eq:kinematics}
\end{equation}

The dynamic model of the robot was exhaustively studied and characterized by Scolari~\cite{Conceicao2007}. 
For a simplified dynamic model, with a reasonable fit to the real robot behavior, it was enough to consider Newton's laws for the dynamics of linear and angular movement~\cite{tipler2007physics}, the dynamics of DC motors and the used speed controllers.

\section{Robot modeling}
\label{sec:robot_modelling}

For starting the modeling process, it was considered the simulation robot designed by Pedro Relvas~\cite{PedroMigueldaSilvaRocha2017} and the measures made by Nascimento \textit{et al.}~\cite{Conceicao2012} (table~\ref{tab:MedicoesRobo}) for a preceding version of this robot. 
Some of the parameters were started with the parameters informed by the manufacturer (table~\ref{tab:caracteristicasMotorCaixaRedutoraEncoders}).

The calibration procedure was divided into two parts.
In the first part, the motor's parameters were estimated, concerning just their independent behavior.
After, the missing parameters, such as the wheels PID gains and the moment of inertia, were estimated using the global behavior of the robot.

\begin{table}[!ht]
    \centering
    \leavevmode
    \caption{Parameter of motors, reduction gear and encoder from their datasheets}
    \begin{tabular}{|c|c|c|} \hline
        \textbf{Description} & \textbf{Value} & \textbf{Unit} \\ \hline
        Reduction gear   &   1:12 & \\ \hline
        Motor resistance ($R_i$) & $0,317$ & $\Omega$ \\ \hline
        Motor electric constant ($K$) & $0,0302$ & N$\cdot$m/A \\ \hline
        Encoder & $256$ & PPR \\ \hline
    \end{tabular}
    \label{tab:caracteristicasMotorCaixaRedutoraEncoders}    
\end{table}

\begin{table}[!ht]
    \centering
    \leavevmode    
    \caption{Starting parameter of the robot for simulation}
    \begin{tabular}{|c|c|c|} \hline
        \textbf{Description} & \textbf{Measure} & \textbf{Unit} \\ \hline
        Robot weight & 26,2 & kg \\ \hline
        Wheels weight & 0,660 & kg \\ \hline
        Wheels radius & 0,0513 & m \\ \hline
        Inertia moment in $x$ axis & 0,629 & kg$\cdot$m$^2$ \\ \hline
        Inertia moment in $y$ axis & 0,658 & kg$\cdot$m$^2$ \\ \hline
        Inertia moment in $z$ axis & 0,705 & kg$\cdot$m$^2$ \\ \hline
    \end{tabular}
    \label{tab:MedicoesRobo}    
\end{table}

\subsection{Motors' dynamical model}
\label{sec:motor_friction}

The case study's robot has three DC brushed motors, whose steady-state voltage-current model is approximated through \eqref{eq:motor_voltage_model}.
In this equation, $u(t)$ and $i(t)$ are the voltage and the current in the motor, and $\omega_m (t)$ is the motor angular speed.
The other parameters are characteristics of the motor, where $R_i$ is the internal resistance, and $K_e$ is the constant of the electromotive force.

\begin{equation}
    u(t) = R_i \cdot i(t) + K_e \cdot \omega_m (t) \label{eq:motor_voltage_model}
\end{equation}

In the same way, the dynamic model of the model can be described through Newton's laws for the rotation movement.
Beside that, it is known that the sum of applied torque is the moment of inertia, $J$, times the angular acceleration, $\frac{\mathrm{d}\omega_m}{\mathrm{d}t}$ \eqref{eq:motor_dynamics_model}. 
In this equation, $K_t$ is the torque constant, and $F_c$ and $B_v$ are the Coulomb and the viscous frictions, respectively.

\begin{equation}
    J \cdot \frac{\mathrm{d}\omega_m}{\mathrm{d}t} = K_t \cdot i(t) - B_v \cdot \omega_m (t) - F_c \label{eq:motor_dynamics_model}
\end{equation}

The equation \eqref{eq:motor_full_model} gives the full model of the robot, in the absence of external forces, relating the two previous equations, considering that $K_e = K_t$ (in the international system of units) and that, in the steady state, there are no acceleration,  $J \cdot \frac{\mathrm{d}\omega_m}{\mathrm{d} t} = 0$.
This equation states a linear relationship in the motor behavior that allows the determination of $B_v$ and $F_c$ through linear approximation.

\begin{equation}
     u(t) = \left(\frac{R_i \cdot B_v}{K_t} + K_t\right) \cdot \omega_m + \frac{R_i \cdot F_c}{K_t} \label{eq:motor_full_model}
\end{equation}

\begin{figure}
    \leavevmode
    \begin{center}        
        \includegraphics[width=\linewidth]{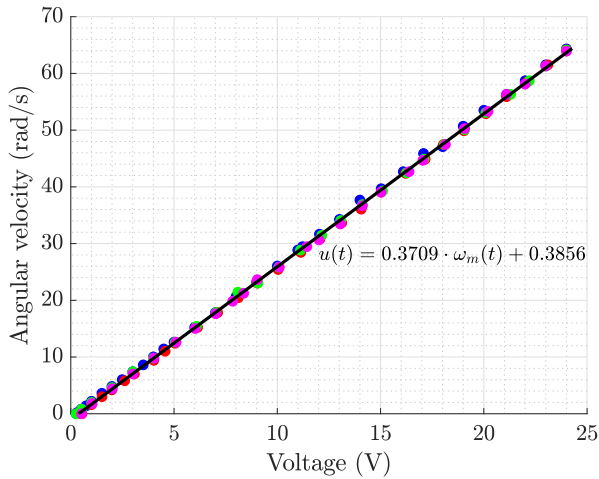}
        \caption{Friction approximation through a linear function, considering the steady-state of the motors under different voltages, with them ``in the air'' (multiple color bullets -- different essays in different wheels).}
        \label{fig:friction_approximation}
    \end{center}
\end{figure}

Once \eqref{eq:motor_full_model} represents the dynamics of the motor in the absence of external forces, to estimate the two friction used in this equation, it should follow a strategy that minimizes the interference of these forces. A typical procedure to do it is to observe the behavior of each motor individually. In this case, the robot was suspended, eliminating the contact of the wheels with the floor. The motor's set corresponds to the set of the motor, gearbox, and wheel. To compute both kinds of frictions, we carried out multiple essays in the different motors with many voltages in the motor's range ($0$ -- $24$ V), measuring the related speed. After that, we searched a rule, through linear approximation, which best fits the dots, as shown in Fig.~\ref{fig:friction_approximation}.
The results show a good approximation for $B_v = 0,0324$~N$\cdot$m/rad/s and $F_c = 0,036735$~N$\cdot$m.

\subsection{Estimation of PID gains and robot's inertia moment}
\label{sec:global_approximation}

Usually, when unknown, due to the complexity of this kind of system, parameters such as the motor's controllers gains, the moment of inertia, or wheels friction are hard to calculate. Because of this, instead of computing them through analytical or experimental methods, iterative computing optimization algorithms tried to approximate the behavior between the real and the simulated robot's response. Different optimization algorithms could be used, but resilient propagation~\cite{Riedmiller1993} or steepest descendant performed well in optimizing a cost function as the minimum square error function \eqref{eq:cost_function}. In this equation, $n$ is the number of sampling points, $\hat{x}_i$ is the simulator's response, to minimize, and $x_i$ is the real robot's response, from the robot essay.

\begin{equation}
    F_c = \frac{1}{n} \cdot \sum^{n}_{i =1} (\hat{x}_i - x_i)^2
    \label{eq:cost_function}
\end{equation}

\begin{figure}[!htb]
    \leavevmode
    \begin{center}        
        \includegraphics[width=\linewidth]{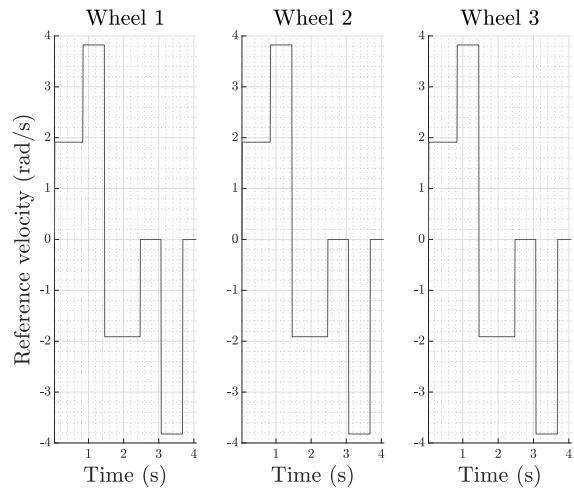}
        \caption{Angular movement excitation model}
        \label{fig:rotation_reference}
    \end{center}
\end{figure}

Due to the high complexity of the robot and its unknown or non-parameterizable features, we attempted to approximate the PID controllers of the simulated robot through a virtual controller, which instead of only approximate the real controllers' behavior, also tries to approximate other features of the robot. Therefore, to compute these virtual PID controllers gains, and the robot's moment of inertia, the resilient propagation (RPROP) algorithm optimized the goal function of \eqref{eq:cost_function}. As an input for the estimation procedure, we considered a pure angular movement excitation signal, as represented in figure~\ref{fig:rotation_reference}. In both cases, the excitation signals were selected to ensure some excitability, without low amplitude velocities, keeping far from motors' dead zone,  and avoiding motors' speed saturation. The duration of each step was selected to ensure that the motors achieve the steady-state, avoiding a long time in this state, maximizing the global time in transient states, the most interesting ones for these issues. Finally, once the essays in the real robot have a restricted area, the total time and the general robot's movement were selected to ensure the robot keeps inside the available area.

Attempting to simplify the approximation method, and avoiding to change wrong parameters, the approximation process was executed in small steps. First of all, the RPROP algorithm searched for the best controller gains, minimizing \eqref{eq:cost_function}, for the virtual PID controllers, considering the response of the real robot to the velocity references of Fig.~\ref{fig:rotation_reference}. Once in this excitation signal, the velocity of the wheels is small, the robot avoids the acceleration limitation effect, optimizing better the controllers' gains. The results reports the good fit reached with this approximation and the controller gains converged to $k_p = 0.06472$, $k_i = 0.043796$ and $k_d = 0$, considering the controller transfer function described in \eqref{eq:controller}.    

\begin{equation}
    C(s) = k_p + \frac{k_i}{s} + k_d \cdot s
    \label{eq:controller}
\end{equation}

\begin{figure}[!htb]
    \leavevmode
    \begin{center}        
        \includegraphics[width=\linewidth]{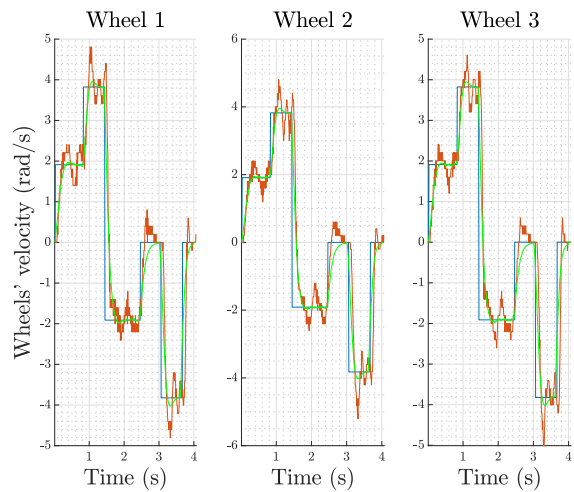}
        \caption{Inertia and virtual PID controllers values optimization to approximate the simulated robot response to the real robot response (blue -- reference, red -- real robot response, green -- simulated robot response).}
        \label{fig:inertia_approximation}
    \end{center}
\end{figure}

In the second iteration, we keep the same velocities reference excitation signal (Fig.~\ref{fig:rotation_reference}) attempting to optimize the moment of inertia in the $z$-axis. Once finished the approximation process, the simulated robot behavior seems as shown in the graphics of Fig.~\ref{fig:inertia_approximation}, with a robot's moment of inertia in the robot's $z$ axis of $1.1$~kg$\cdot$m$^2$.

\section{Results}
\label{sec:results}

Although the results in the section~\ref{sec:robot_modelling} show that there is proper general fitting for the reached results, an extra validation of them is still essential.

\begin{figure}[!htb]
    \leavevmode
    \begin{center}        
        \includegraphics[width=\linewidth]{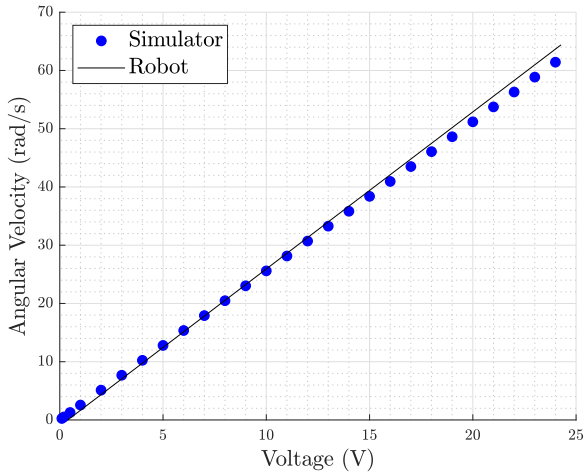}
        \caption{Validation of the wheels' friction model}
        \label{fig:friction_validation}
    \end{center}
\end{figure}

Concerning the estimation of the wheels' friction, we used a standard method through linear approximation. The response illustrated in Fig.~\ref{fig:friction_validation} validates that approximating a set of samples that relates the wheel's velocity to the applied voltage through a linear function to estimate Coulomb and viscous frictions is a good strategy. However, it is possible to observe a slight deviation between the estimated wheels' model in the section~\ref{sec:motor_friction} (Fig.~\ref{fig:friction_approximation}) and the response got from the simulated robot,  due to non-modeled and unpredictable phenomena in the real robots, such as the floor imperfections, wheels imperfections, slippage, and others.

Moreover, despite the results reached in the responses of Fig.~\ref{fig:inertia_approximation} do not fit equally on the robot behavior, it is possible to consider that we reached a good approximation of the robot to simulation. Extra noise and oscillations could be related to non-modelable features such as the slippage.

\begin{figure}[!htb]
    \leavevmode
    \begin{center}        
        \includegraphics[width=\linewidth]{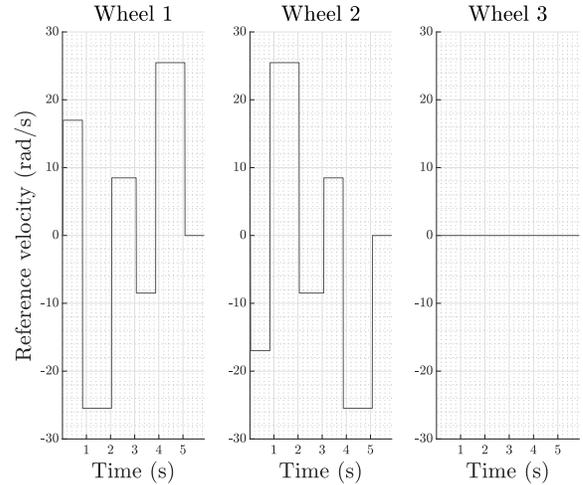}
        \caption{Linear movement excitation model in the $\vectorsym{v}$ robot axis}
        \label{fig:linear_reference}
    \end{center}
\end{figure}

For validating the reached results, it was considered a second excitation model with linear velocity in the robot $\vectorsym{v}$ component (Fig.~\ref{fig:linear_reference}).
Differently of the excitation model considered previously (Fig.~\ref{fig:rotation_reference}), the model keeps the robot movement near the saturation (avoiding it), which becomes visible since the maximum acceleration limitation configured to $61.09$~rad/s$^2$.

\begin{figure}[!htb]
    \leavevmode
    \begin{center}        
        \includegraphics[width=\linewidth]{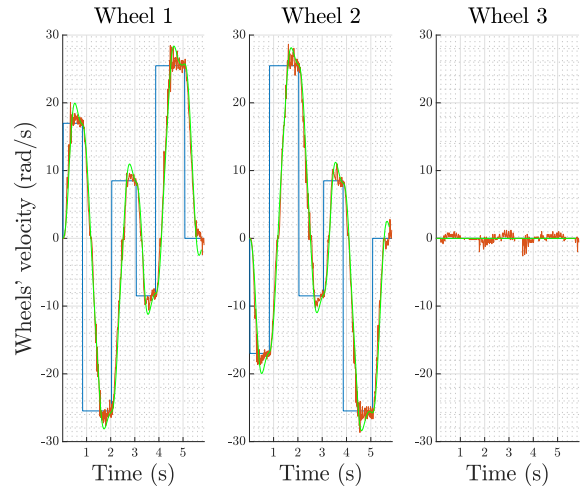}
        \caption{Validation of the simulated robot response using a linear front and back movement (blue -- reference, red -- real robot response, green -- simulated robot response).}
        \label{fig:linear_approximation}
    \end{center}
\end{figure}

Because of wheel slip problems, the real robot was limited to acceleration, and the simulated robot's response reaches a very close fit to the real ones (Fig.~\ref{fig:linear_approximation}) even with this limitation.

\section{Conclusions}
\label{sec:conclusions}

This work attempts to estimate a general robot model of a robot of the 5DPO robotics soccer team for SimTwo simulator software~\cite{SimTwo}. The robot parameters were initialized with the computed value for previous versions of this robot, due to their similarities. Then a first-order system approached the motor and wheel assembly through linear approximation. Finally, an iterative algorithm approximates the global robot behavior using a sample of the response of the real robot to an excitation model, optimizing the motor's velocity controller gains and robot's moment of inertia.

Although the good results reached in this robot modeling strategy, extensive validation of the simulation robot needs to be done, using other excitation signals and movement variability. These tests should contemplate other movements of the robot, such as lateral displacements and mixed displacements and rotations in all directions.

%
%
%
 \bibliographystyle{IEEEtran}
 \bibliography{myrefs}

\begin{thebibliography}{10}
\providecommand{\url}[1]{#1}
\csname url@samestyle\endcsname
\providecommand{\newblock}{\relax}
\providecommand{\bibinfo}[2]{#2}
\providecommand{\BIBentrySTDinterwordspacing}{\spaceskip=0pt\relax}
\providecommand{\BIBentryALTinterwordstretchfactor}{4}
\providecommand{\BIBentryALTinterwordspacing}{\spaceskip=\fontdimen2\font plus
\BIBentryALTinterwordstretchfactor\fontdimen3\font minus
  \fontdimen4\font\relax}
\providecommand{\BIBforeignlanguage}[2]{{%
\expandafter\ifx\csname l@#1\endcsname\relax
\typeout{** WARNING: IEEEtran.bst: No hyphenation pattern has been}%
\typeout{** loaded for the language `#1'. Using the pattern for}%
\typeout{** the default language instead.}%
\else
\language=\csname l@#1\endcsname
\fi
#2}}
\providecommand{\BIBdecl}{\relax}
\BIBdecl

\bibitem{siegwart}
R.~Siegwart, I.~R. Nourbakhsh, and D.~Scaramuzza, \emph{Introduction to
  autonomous mobile robots}.\hskip 1em plus 0.5em minus 0.4em\relax MIT Press,
  2011.

\bibitem{Tang}
{Jun Tang}, K.~{Watanabe}, and Y.~{Shiraishi}, ``Design and traveling
  experiment of an omnidirectional holonomic mobile robot,'' in
  \emph{Proceedings of IEEE/RSJ International Conference on Intelligent Robots
  and Systems. IROS '96}, vol.~1, Nov 1996, pp. 66--73 vol.1.

\bibitem{ZLAJPAH}
L.~Žlajpah, ``Simulation in robotics,'' \emph{Mathematics and Computers in
  Simulation}, vol.~79, no.~4, pp. 879 -- 897, 2008, 5th Vienna International
  Conference on Mathematical Modelling/Workshop on Scientific Computing in
  Electronic Engineering of the 2006 International Conference on Computational
  Science/Structural Dynamical Systems: Computational Aspects.

\bibitem{Yuan}
Y.~xiang Yuan, ``A new stepsize for the steepest descendent method,''
  \emph{Journal of Computational Mathematics}, vol.~24, no.~2, pp. 149--156,
  2006.

\bibitem{Riedmiller1993}
M.~Riedmiller and H.~Braun, ``{A direct adaptive method for faster
  backpropagation learning: The RPROP algorithm},'' \emph{IEEE International
  Conference on Neural Networks - Conference Proceedings}, vol. 1993-Janua, pp.
  586--591, 1993.

\bibitem{5dpo}
T.~Nascimento, M.~Pinto, H.~Sobreira, F.~Guedes, A.~Castro, P.~Malheiros,
  A.~Pinto, H.~Alves, M.~Ferreira, P.~Costa \emph{et~al.}, ``5dpo 2011: Team
  description paper,'' \emph{no. Robocup, Janeiro}, 2011.

\bibitem{Caldas2009}
\BIBentryALTinterwordspacing
R.~M. D.~S. Caldas, ``{Modela{\c{c}}{\~{a}}o e Simula{\c{c}}{\~{a}}o de um
  Robot Omnidireccional de 3 rodas},'' Tese de mestrado, Universidade do Porto,
  2009. [Online]. Available:
  \url{https://repositorio-aberto.up.pt/handle/10216/59099}
\BIBentrySTDinterwordspacing

\bibitem{Conceicao2007}
\BIBentryALTinterwordspacing
A.~G.~S. Concei{\c{c}}{\~{a}}o, ``\BIBforeignlanguage{Portuguese}{{Controlo e
  coopera{\c{c}}{\~{a}}o de rob{\^{o}}s m{\'{o}}veis aut{\'{o}}nomos
  omnidireccionais}},'' Phd Thesis, University of Porto, 2007. [Online].
  Available: \url{http://hdl.handle.net/10216/11398}
\BIBentrySTDinterwordspacing

\bibitem{Nascimento2009}
T.~P.~D. Nascimento, ``{Controle de trajet{\'{o}}ria de rob{\^{o}}s
  m{\'{o}}veis omni-direcionais : uma abordagem multivari{\'{a}}vel},'' Ph.D.
  dissertation, Universidade Federal da Bahia, 2009.

\bibitem{Conceicao2006}
\BIBentryALTinterwordspacing
A.~S. Concei{\c{c}}{\~{a}}o, A.~P. Moreira, and P.~Costa, ``{Controller
  optimization and modelling of an omni-directional mobile robot},'' in
  \emph{CONTROLO 2006}, 2006. [Online]. Available:
  \url{http://hdl.handle.net/10216/79811}
\BIBentrySTDinterwordspacing

\bibitem{Conceicao2009}
\BIBentryALTinterwordspacing
A.~S. Conceicao, A.~P. Moreira, and P.~J. Costa, ``{Practical approach of
  modeling and parameters estimation for omnidirectional mobile robots},''
  \emph{Mechatronics, IEEE/ASME Transactions on}, vol.~14, no.~3, pp. 377--381,
  jun 2009. [Online]. Available:
  \url{http://ieeexplore.ieee.org/xpls/abs{\_}all.jsp?arnumber=4797830}
\BIBentrySTDinterwordspacing

\bibitem{SimTwo}
C.~Paulo, G.~José, L.~José, and M.~Paulo, ``Simtwo realistic simulator: A
  tool for the development and validation of robot software,'' \emph{Theory and
  Applications of Mathematics \& Computer Science}, vol.~1, no.~1, pp. 17--33,
  2011.

\bibitem{gazebo}
N.~{Koenig} and A.~{Howard}, ``Design and use paradigms for gazebo, an
  open-source multi-robot simulator,'' in \emph{2004 IEEE/RSJ International
  Conference on Intelligent Robots and Systems (IROS) (IEEE Cat.
  No.04CH37566)}, vol.~3, Sep. 2004, pp. 2149--2154 vol.3.

\bibitem{vrep}
M.~F. E.~Rohmer, S. P. N.~Singh, ``Coppeliasim (formerly v-rep): a versatile
  and scalable robot simulation framework,'' in \emph{Proc. of The
  International Conference on Intelligent Robots and Systems (IROS)}, 2013,
  www.coppeliarobotics.com.

\bibitem{Smith2007}
\BIBentryALTinterwordspacing
R.~Smith, ``{Open Dynamics Engine},'' 2007. [Online]. Available:
  \url{http://www.ode.org}
\BIBentrySTDinterwordspacing

\bibitem{tipler2007physics}
P.~Tipler and G.~Mosca, \emph{Physics for Scientists and Engineers}, ser.
  Physics for Scientists and Engineers: Standard.\hskip 1em plus 0.5em minus
  0.4em\relax W. H. Freeman, 2007.

\bibitem{PedroMigueldaSilvaRocha2017}
\BIBentryALTinterwordspacing
R.~{Pedro Miguel da Silva Rocha}, ``{Fus{\~{a}}o sensorial e
  coopera{\c{c}}{\~{a}}o em equipas de rob{\^{o}}s m{\'{o}}veis},'' Mestrado,
  Universidade do Porto, 2017. [Online]. Available:
  \url{http://hdl.handle.net/10216/105366}
\BIBentrySTDinterwordspacing

\bibitem{Conceicao2012}
\BIBentryALTinterwordspacing
A.~G. Concei{\c{c}}{\~{a}}o, A.~Moreira, P.~Costa, P.~Costa, and T.~Nascimento,
  ``{Modeling omnidirectional mobile robots: an approach using SimTwo},'' in
  \emph{Controlo'2012}, Porto, jul 2012, p.~6. [Online]. Available:
  \url{http://www.apca.pt/index.php/site/pubdetails/24}
\BIBentrySTDinterwordspacing

\end{thebibliography}

\end{document}